\documentclass[12pt]{article}
\usepackage[utf8]{inputenc}
\usepackage{hyperref}
\usepackage{float}
\usepackage{graphicx}
\graphicspath{ {./images/} }

\begin{document}
\begin{center}
\textbf{CONVOLUTIONAL NEURAL NETWORK IN STOCK PRICE MOVEMENT PREDICTION}\\
\vspace{5mm}
\textbf{\textit{Kunal Bhardwaj}}\\
Department of Mathematics, Shiv Nadar University\\
kb702@snu.edu.in\\
\end{center}


%
\begin{abstract}
  The usability of Neural Networks in different sectors has been a hot topic of research for many years. With technological advancements and exponential growth of data, we have been unfolding different capabilities of neural networks in different sectors. In this paper, I have tried to use CONVOLUTIONAL NEURAL NETWORK in STOCK PRICE MOVEMENT PREDICTION. In other words, I have tried to construct and train a convolutional neural network on past stock prices data and then tried to predict the movement of stock price i.e. whether stock price would rise or it would fall, in the coming time.
\end{abstract}

\section{Introduction}
I have constructed a convolutional neural network for 1-D time series data of past stock prices and then checked its performance on testing data. I have used the past stock prices of NIFTY 50 to train and test my network.\\
For my dataset, I have chosen the period from April 2013 to September 2016 for the minute wise data of stock prices and then divided this dataset into training and testing sets accordingly.\\
After obtaining the final results on testing data, I have evaluated precision, accuracy, recall and F1 score: evaluation metrics, to get the efficacy of my model.\\
Now, before getting started with the procedure of constructing my model, I would like to give an overview of functioning of neural networks and convolutional neural networks so that it would be helpful for readers to relate the terms.\\

Let's start with Neural Networks first:
\begin{itemize}
    \item \textbf{Neural Network}:A neural network(or more precisely an artificial neural network) is a network of interconnected layers consisting nodes which operates to recognize underlying relationships in a set of data. It mimics the way in which the human brain operates to process information.\\
 A neural network contains layers of interconnected nodes as follows:\
    \begin{itemize}
        \item  The first layer is known as "Input layer" and this layer contains inputs or features used for training a model.
        \item The last layer is known as "Output layer" and this layer shows the final outcomes of a model.
        \item The layers in between input and output layers are known as "Hidden layers" and these are the layers where the main processing of information takes place.
    \end{itemize}
    \textit{Note} - There are no nodes or neurons in the input layer, it contains only the inputs.\\
    Hidden layers and the output layer of a neural network consists nodes(neurons) and the number of nodes in a layer and the number of hidden layers between input and output layers are problem specific and vary from problem to problem.\\
    The data enters the neural network through the input layer.\\
    The layers of a neural network where all the inputs from one layer are connected to every neuron of the next layer are known as "fully connected layers".\\
    
    The processing of information from the input layer to other layers takes place in the following way:\\
    
    Each input of the input layer is passed to all the nodes of subsequent hidden layer in this way: firstly, at every node of the subsequent layer, weights are assigned to all inputs and linear combination of inputs with corresponding weights is calculated and a bias term is added to the calculated result (for inputs, weights and biases at different nodes in a layer need not to be the same) and then, the calculated value is passed to a non-linear activation function and output is calculated.\\
    These calculated outputs of activation functions are stored at values at each node and then these values at different nodes become inputs for next layer and is passed to every node of the next layer in a similar way and in this manner, the information is processed through all the layers of a network and after obtaining final results at the output layer, predicted results are compared with the actual results and based on that, loss function is calculated.\\
    This way of processing the information in a forward manner from the input layer to the output layer is known as "forward propagation" in a neural network.
    Our goal is to find those values of weights and biases for each node so that the loss function's value can be minimized.\\
    Once computing the loss function at the output layer, this is propagated to all the hidden layers of the network in backward direction and an optimizer(e.g. gradient descent) is used at each layer starting from the output layer to all the hidden layers, to find the optimal values of weights and biases for which the loss function's value can be minimized and this way of processing the information from output layer to previous layers in a backward direction is known as "back propagation" in a neural network.\\
    A number of iterations of this forward and back propagation of information is carried out in order to achieve the optimal values of weights and biases in a neural network and to make a neural network ready to be able to perform on unseen data.\\
    
    Now, let's start with Convolutional Neural Network:
    \item\textbf{Convolutional Neural Network}: A Convolutional Neural Network(CNN) is a deep learning algorithm which is basically a network of mainly two types of layers:CONV+POOL layers and Fully connected layers.\\
    
    \textit{Note:} Because of the remarkable property of CNN to extract spatial features from data, CNN models are highly used in image analysis tasks e.g. taking images data(2-D or 3-D) as input and then train CNN models on that data to extract spatial features and enable output prediction, but this does not imply that we can not use CNN on 1-D data.\\
    
     In general, when we have 1-D data as input, like in our case where we have time-series data which is 1-D only, there are two ways to use CNN on that data, one way is to convert 1D-input into a 2D or 3D plot by adding additional information to data and then use it as an image input for a CNN to train the model; while the other way is to take advantage of CONV1D function of tensorflow in order to be able to perform operation of convolution on 1-D data itself. I have used the latter one.\\
     \begin{itemize}
         \item In CNN, filter (also called as kernel) is used that is basically an array of values, these values are called as weights. The dimension of a filter is known as kernel size, it is generally less than the dimension of the input.\\
         In CNN, the value of a node in the layer subsequent to the input layer is obtained in the following manner: The filter slides over the input starting from the very first input value and each time, convolution operation(that is basically a linear combination of weight coefficients of the filter and corresponding input values inside the patch of input) is performed between filter and the patch of input which is of the same size as of kernel size. To the calculated value of convolution, a bias term is added then the resulting output is passed to a non-linear activation function and output is calculated.\\
         During each slide of filter, the value of activation function that is calculated becomes the value of a corresponding node in the subsequent layer.\\
         Multiple filters are used to extract different features from the input.\\
         In CNN, the output of an activation function(or the outputs stacked together, in case of multiple filters) is called as "feature map".\\
         Pooling operation is generally performed on the obtained feature map which reduces the dimensionality of feature map by retaining important information to speed up the computation. These layers where convolution and pooling operations are performed are called as "CONV+POOL" layers together.\\
         The resulting output from a pooling layer then becomes the input for the next convolutional layer and all operations are performed again similarly; in this way, information is processed through other CONV+POOL layers also; later, the output of the final pooling layer is flattened and connected with a network of fully connected layers and then information is processed as we have seen in the previous section and the final outcome is predicted at the output layer.\\
         Then, at the output layer, loss function is computed by comparing actual and predicted values and is propagated in backward direction through all the convolutional and hidden layers and at each layer(except pooling layers since pooling layers just reduce the dimensions of the feature maps, they don't contain any parameters to be computed), an optimizer is used to find the optimal values of weights and biases for which the value of loss function can be minimized. And as, we have seen in previous section, this is achieved through a number of iterations of forward and back propagation.\\
         \end{itemize}
         
         \textbf{\textit{Note:}} Non-linear activation functions are used in neural networks because of the following reasons:
         \begin{itemize}
             \item They allow back propagation since they have a derivative function which is related to the inputs.
             \item They allow stacking of multiple layers of nodes to create a deep neural network,multiple hidden layers are used to learn complexities of a dataset with high levels of accuracy.
             \item They also helps in solving the problem of gradient vanishing.
         \end{itemize} 
     
\end{itemize}

\section{Data Pre-processing}
To train any neural network, data is required; so my first job was to get the dataset and then extract the important features from it which would be used to train my neural network.\\
After pre-processing my original dataset of NIFTY50's minute wise stock prices for the period of April 2013 to September 2016, I used it for training and testing my model.\\
The ordered sequence of steps that were taken for pre-processing the dataset are given below:

\subsection{Pre-processing of Data}
Data pre-processing is necessary because the original dataset may contains a lot of noise or irrelevant information which is not required for training a model; so, to decrease the computation time and also to avoid any interference of irrelevant information with the final result of a model, data pre-processing is required as a starting step.

\subsubsection{\textbf{Adding required information to the dataset}}
First of all, I dropped those rows from my dataset which contained any missing value or any non-numerical value.
Initially, my dataset had seven columns(features): 'Open', 'High', 'Low', 'Close', 'Date', 'Time', 'Index'; and my goal was to do the binary classification i.e. whether my model would be able to predict correctly when stock price would go up and down or not, I wanted two labels in my dataset, one label to show that the stock price would rise after 15 minutes and other label to show that it would fall after 15 minutes.\\
But my original dataset did not have any such column which could indicate labels. So, I inserted two columns named 'High15' and 'Label' in my dataset.\\
I used fifteen minutes window i.e to label a row that was at the ith position, I used the data of the row which was at i+15th position in the following manner:\\ The value under 'High15' column for a row which was at ith index was actually the stock price under 'High' column of the row which was at i+15th index and to calculate the label for a row, its values under 'High' column and 'High15' column were compared and if 'High15' price was greater than 'High' price then that row was labelled as '1' otherwise '0'.\\
Label '1' was used to indicate that High price of a stock would go up after fifteen minutes and label '0' was used to indicate that it would go down after fifteen minutes.
\subsubsection{\textbf{Cleaning of Dataset: Removing unnecessary information from the dataset}}
After adding required labels in my dataset, it consisted nine columns(features): 'Open', 'High', 'Low', 'Close', 'Date', 'Time', 'Index', 'High15' and 'Label'; where 'Date', 'Time', 'Index' columns did not contain any useful information which could be used to train my model, so I decided to drop these columns. And also, I did not want the final predictions of my model to be influenced by the information of future stock prices because my main task was to do future predictions and it did not make sense to use future information in order to predict it, so I dropped 'High15' column also and kept only first four columns(features) in my dataset which were: 'Open', 'High', 'Low', 'Close'.\\
Among these four selected columns, I decided to predict the movement of stock prices under 'High' column after 15 minutes i.e. tried to predict when the High prices of stocks would rise after 15 minutes and when it would fall after 15 minutes.\\

\subsubsection{\textbf{Data Separation and Normalization of Training and Testing Data}}
I divided my dataset into two parts, I used 70 percent of dataset for training my model and reserved remaining 30 percent of it for testing the performance of my model.
\begin{figure}[H]
    \centering
    \includegraphics[scale=.8]{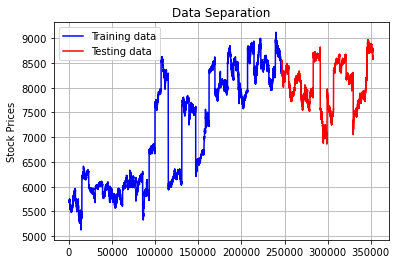}
    \caption{Separation of data into training and testing data}
    \label{fig:1}
\end{figure}
But there was significant difference between the values of a column, so, I adopted "min-max normalization" technique to normalize the values of a column and to keep each value in the range of 0-1.\\
In a column, if $x$ denotes the value of a feature then after normalizing it, this value will be converted into $x'$ by using this formula under min-max normalization:
\begin{equation}
x' = \frac{x - min}{max - min}
\end{equation}
where max denotes the maximum value and min denotes the minimum value of that feature, across that column.\\ 

\section{Network Engineering}
\subsection{Convolution on 1-D time series data}
As my dataset consisted of one dimensional time-series data, I decided to use Conv1D function of tensorflow to build my network.\\
In Conv1D, filter slides only along one dimension, it performs convolution operation on input data and generates output. The output of this function is basically a n-dimensional array where the value of n is decided by the number of filters in a convolutional layer.\\
\subsection{Building CNN for stock price movement prediction}
\subsubsection{\textbf{Layers and Activation functions}}
As I have explained earlier, a convolutional neural network consists of CONV+POOL layers and fully connected layers, I constructed a network of total five layers where first three were convolutional layers and later two were fully connected layers.\\
In my network, I adopted "same" padding to pad the inputs for each subsequent convolutional layer in order to prevent the reduction in dimension of outputs after convolution operation and the kernel size for filters in each convolutional layer was set to three.\\
Now, since non-linear activation functions are used in a neural network, I used two different activation functions in my network named as ReLU and LeakyReLU, which are defined as follows:\\
\begin{itemize}
    \item \textbf{ReLU}: This function returns zero if its input is less than or equal to zero and returns the same input as output if input is greater than zero.
    \item \textbf{LeakyReLU}: This function returns the same input as output when it is greater than zero but when its input is less than or equal to zero then it returns $\alpha$ times input as output, where $\alpha$ is a hyper-parameter with a very small value, generally 0.001.  
\end{itemize}
I had to use two activation functions instead of only one because, however, ReLU function is a useful activation function to solve the problem of gradient vanishing, but the problem would still exist when the information includes negative values.\\
Thus, for the first convolutional layer, I chose ReLU as my activation function and for other layers, I chose LeakyReLU as my activation function.\vspace{5mm}

Then, I used "Max-pooling" technique for one dimensional data to reduce the dimensions of feature maps by retaining important information which facilitates shorter computation time.\\
I obtained two pooling layers, using Max-pooling and a kernel of size 2, one with second convolutional layer and other with third convolutional layer.\\  

After that, since my ultimate task was to do the binary classification, for the output of the last fully connected layer(fifth layer), I adopted "Sigmoid" function as my activation function which returns the probabilities of the output belonging to different classes (where the bigger probability of belonging to either class decides the label).\\
For an input x, sigmoid function's output(f(x)) is defined as:\\
\begin{equation}
      f(x) = \frac{1}{1+e^{-x}}
\end{equation}\\

\subsubsection{\textbf{Quantifying Loss}}
After obtaining the final outcome at the output layer, my job was to compute the loss function J(W) where W stands for weight vector. I used "binary cross-entropy" loss function to compute the loss, binary cross-entropy function uses the following formula to compare predicted values with the actual values:\\
\begin{equation}
      J(W) = \frac{1}{n}\sum\limits_{i=1}^n -(y_i * log(\hat{y_i})) + (1-y_i) * log(1 - \hat{y_i}))
\end{equation}\\
where n is the total number of samples in the training set, $\hat{y_i}$ is the predicted probability that output of ith sample belongs to label 1, 1 - $\hat{y_i}$ is the predicted probability that output of ith sample belongs to label 0, $y_i$ is the actual label for ith sample which takes either 1 or 0.\\

When the observation actually belongs to label 1, the first part of the formula becomes active and the second part vanishes and vice versa in case when the actual label is 0.\\

\subsubsection{\textbf{Minimizing Loss}}

To minimize the loss, instead of using gradient descent technique, I adopted "Adaptive Moment Estimation(Adam)" optimizer which basically works on the principle of gradient descent technique but in a more advanced way that solves the problem of gradient vanishing and enables the model to find the optimal values of weight coefficients.\\

Now, the question arises, why is the gradient descent technique not sufficient to find the optimal values of weight coefficients?\\
To answer this, I would like to introduce two shortfalls of using  gradient descent technique as an optimizer in a network:
\begin{enumerate}
    \item Gradient descent gets stuck at local minima.
    \item The learning rate does not change in gradient descent.\\
\end{enumerate}

These shortfalls restrain a user to simply adopt gradient descent as an optimizer and to overcome these shortfalls one by one, the following methods are used:\\
\begin{itemize}
    \item To deal with problem 1, a technique known as "Stochastic Gradient Descent with Momentum" is used. It uses the weighted sum of the past gradients for updating weights in the following manner:\\
    It calculates $m_t$ s.t.\\
    \begin{equation}
      m_t = \frac{\beta}{m_{t-1}} + (1 - \beta)\frac{\partial J(W)}{\partial w_t}
    \end{equation}\\
    where $\frac{\partial J(W)}{\partial w_t}$ gives current gradient at time t, J(W) denotes the loss function, $m_{t-1}$ gives previously accumulated gradient till time t-1.\\
    $\beta$ gives the value of how much weightage to be given to the current gradient and previously accumulated gradient. Generally, 10 percent weightage is given to the current gradient and 90 percent is given to the previously accumulated gradient.\\
    
    After computing, $m_t$, stochastic gradient descent with momentum updates the weight vector by using the following formula:\\
     \begin{equation}
      w_{t+1} = w_t - \alpha m_t
    \end{equation}
    where $\alpha$ denotes learning rate.\\
    
    This weighted sum of past gradients will provide the optimizer required push when it would stuck at local minima and enable the optimizer to find better values of weight coefficients in comparison with just gradient descent.\\
    
    \item To deal with problem 2., a technique known as "RMSprop" is used. It uses the weighted sum of the squared past gradients and changes the learning rate according to the loss function in the following manner\\
    \begin{equation}
      w_{t+1} = w_t - \frac{\alpha}{\sqrt{s_t }+ \epsilon}.\frac{\partial J(W)}{\partial w_t}\\
    \end{equation}
    where\\
     $s_t = \beta s_{t-1} + (1 - \beta) (\frac{\partial J(W)}{\partial w_t})^2 $ \\
     
     \textit{Note} - Square of the gradients is taken to eliminate the effect as some gradients may be +ve and some may be -ve, $\epsilon$ is the error term which is added so that denominator does not become zero; generally, very small value of $\epsilon$ is taken.\\ 
     
     This denominator under the learning rate ($\alpha$) changes $\alpha$ according to the loss function that leads to faster convergence to the optimal solution.\\

\item In precise terms, the Adam optimizer is a blend of Stochastic Gradient Descent with Momentum and RMSprop techniques which allows it to minimize the loss function more efficiently in comparison with gradient descent.\\

The weight update equation for Adam becomes:\\
\begin{equation}
      w_{t+1} = w_t - \frac{\alpha}{\sqrt{s_t} + \epsilon}m_t\\
\end{equation}
\end{itemize}

\subsubsection{\textbf{Problem of overfitting and the optimal number of epochs}}
\textbf{Epoch}: One epoch is when an entire dataset is passed forward and backward through the neural network once.\\
To set the number of epochs in such a manner so that the achievable good value of accuracy on testing data would be achieved in shorter time and simultaneously, the chances of overfitting would get reduced, I used a technique known as "Early stopping", I used early stopping with callbacks and set the upper bar on the number of epochs at 25.\\
I set the value of patience to 5 in callbacks, callbacks restrain a network to run for more number of epochs, than the set value of patience, with no improvement which saves a lot of computation time.\\

I also adopted the "Dropout" technique to reduce the chances of overfitting and to promote generalization in my network. Dropout technique randomly set some nodes to 0 in each iteration during training.\\

Thus, I used both dropout and early stopping techniques in my network to prevent the problem of overfitting.

\subsection{Parameters}
Different parameters in my network can be concluded as follows:\\
\begin{itemize}
    \item \textbf{Number of layers:}\\
    There are five layers in my model among which there are three convolutional layers and two are fully connected layers.\\
    
    Setting optimal number of layers is a crucial step while building a neural network, large number of convolutional layers may increase the complexity of the network and also may lead to the problem of gradient vanishing while small number of convolutional layers may make the model unreliable; that's why to balance the computation speed and efficacy of my model, I decided to choose five layers.\\
    
    \textit{Note}- The pooling layers in my network are not counted separately, I have included them in my count with convolutional layers.\\
    \item \textbf{Number of filters and kernel sizes:}\\
    The number of filters in each convolutional layer is specified as follows:
    \begin{itemize}
        \item First convolutional layer : 32
        \item Second convolutional layer: 64
        \item Third convolutional layer : 128
    \end{itemize}
    And there are two fully connected layers in my network, first one with 128 nodes and other with 256 nodes.\\
    The size of kernel for all filters is 3.\\
    \item \textbf{Batch size}\\
    In general, it is not recommended to pass the entire dataset into a neural network in one go when the available dataset is large because it may decrease the computation speed and also may affect the efficiency of a model. So, to overcome this, dataset is divided into certain batches and then network access the dataset batch by batch only which results in faster computation and a more efficient model.\\ 
    \textbf{Batch size} refers to the number of samples present in one batch. It is carefully chosen because too small batch size has the risk of making learning too stochastic and will converge to unreliable models while too big batch size would not fit into memory and increase the computation time.\\
    Considering all these aspects, I decided to choose 1000 samples in my one batch.
\end{itemize}
\begin{figure}[H]
    \begin{center}
    \includegraphics[scale=.8]{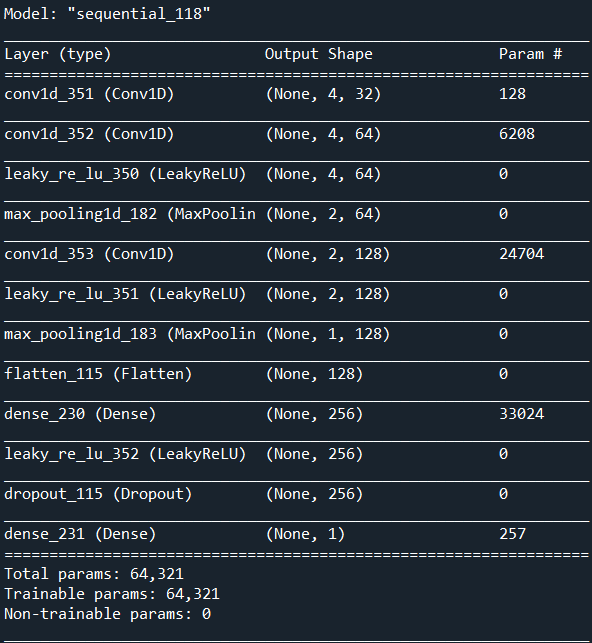}
    \caption{Architecture of my model}
    \label{fig:2}
    \end{center}
\end{figure}
\section{Findings}
To evaluate the performance of my model, I have used these four different evaluation metrics.\\

Let me first introduce the meaning of terms which are used in formulation of evaluation metrics:
\begin{itemize}
    \item TP = True Positive i.e. correctly assigned as 1.
    \item FP = False Positive i.e. wrongly assigned as 1.
    \item TN = True Negative i.e. correctly assigned as 0.
    \item FN = False Negative i.e. wrongly assigned as 0.
\end{itemize}
\begin{enumerate}
    

    \item \textbf{Accuracy :} Accuracy is the most common evaluation metric for classification models because of its simplicity and interpretation. It indicates closeness of the measurements to a specific value. It is calculated by using the formula given below:\\
    \begin{equation}
      Accuracy = \frac{TP + TN}{TP + FP + TN + FN}
    \end{equation}\\
    
    \item \textbf{Precision :} Precision is the proportion of true positive on all positives predictions. A precision of 1 means that you have no false positive. It is calculated as per the formula given below:\\
    \begin{equation}
      Precision = \frac{TP}{TP + FP}
    \end{equation}\\
    
    \item \textbf{Recall :} Recall is the proportion of true positives on all actual positive elements. A recall of 1 means that you have no false negative. The formula for calculating recall is given below:\\
    \begin{equation}
      Recall = \frac{TP}{TP + FN}
    \end{equation}\\
    
    \item \textbf{F1-score :} F1-score combines precision and recall relative to a specific positive class. It can be interpreted as the weighted average of the precision and recall and it is calculated by using the formula given below:
    \begin{equation}
      F1-score = \frac{2 * Precision * Recall}{Precision + Recall}\\
    \end{equation}
    
    \textit{Note} - Each evaluation metric defined above reaches its best value at 1 and worst value at 0.\\
    \end{enumerate}
    
    \begin{itemize}
    \item Along with my main dataset of NIFTY50 stock prices, I have also run my model a number of times on different datasets of these following companies:\\
    'HDFC BANK', 'ADANI PORTS', 'ICICI BANK' and 'BAJAJ FINANCE'.\\
    
    Due to the limited accessibility to the minute-wise data of stock prices; I decided to use the day-wise data available for stock prices of these companies. I tried to fit my model on the available datasets using the same four features of stock price for these companies: 'Open', 'High', 'Close' and 'Low'. I tried to predict whether the 'High' prices of stocks for these companies would rise or fall after 15 days and then checked my model's performance using evaluation metrics.\\
    
    \item In order to get more number of training samples as well as more data for testing the performance of my model, I used day-wise stock prices data for past twenty years, for all these companies.
    \end{itemize}
    
     I calculated evaluation metrics multiple times on each dataset and every time, I was able to achieve more than 0.5 value for accuracy on both training and testing data, more than 0.95 value for recall and more than 0.5 value for precision and more than 0.65 value for F1-score.\\
     
     \textbf{Remarks:} As far as, I could interpret the obtained results of evaluation metrics, the reason for not obtaining higher values of accuracy, precision and F1-score is that if I go by the formulae of calculating these evaluation metrics then it's observable that my model is not able to capture the trends required to predict the fall in the stock price in future, that's why the number of True Negatives and even False Negatives got decreased in the final outcomes resulting in a high value of Recall and the number of False Positives got increased which eventually affected the values of Accuracy, Precision and F1-score in that way.
     \begin{figure}[H]
         \centering
         \includegraphics[scale=.9]{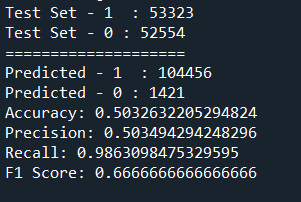}
         \caption{Values of evaluation metrics for a randomly selected run}
         \label{fig:3}
     \end{figure}
     
 \subsection{Performance Plots}
 On that same randomly selected run for which the values of evaluation metrics shown in previous section, to visualize the performance of my model, plots for loss function's values and accuracy on both training and testing data were obtained as follows:
 \begin{figure}[H]
     \centering
     \includegraphics[scale=.66]{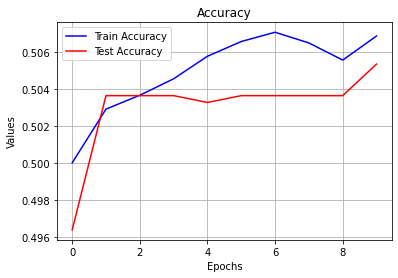}
     \caption{Accuracy plot for that same randomly selected run}
     \label{fig:4}
 \end{figure}
 
 \begin{figure}[H]
     \centering
     \includegraphics[scale=.66]{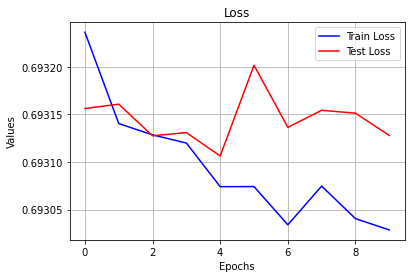}
     \caption{Loss plot for that same randomly selected run}
     \label{fig:5}
 \end{figure}
 
Generally, while training a neural network it is observed that when we increase the number of iterations then a network would be able to capture more trends in the dataset but the reason why I decided to remain stick to my upper bar of 25 iterations along with early stopping for training my model can be well understood through the following plots where I set my number of iterations to 300 and did not use early stopping technique in my network:
\begin{figure}[H]
    \centering
    \includegraphics[scale=.622]{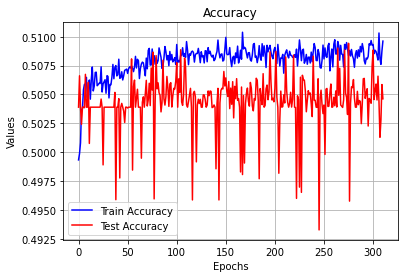}
    \caption{Accuracy plot for a large number of epochs (iterations), for both training and testing data}
    \label{fig:6}
\end{figure}
\begin{figure}[H]
    \centering
    \includegraphics[width=.65\textwidth]{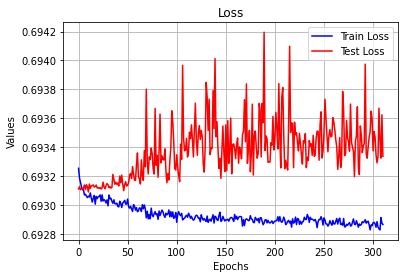}
    \caption{Loss plot for a large number of epochs (iterations), for both training and testing data }
    \label{fig:7}
\end{figure}

 It is easily observable from the plot of Loss function that the value of loss function would keep decreasing for training data as the number of iterations increases but for testing data, this value would decrease for some initial iterations only, after that it would increase with increment in number of iterations; that is a typical case of overfitting, that's why I used early stopping to avoid this kind of situation and used less iterations to train my model.

\section{Conclusion}
In this paper, I have constructed a CNN model to predict the stock prices movement. I have used 1-D time series data as my input to the model, Conv1D function of tensorflow to learn and extract useful features from the data, Max-pooling to save computation time which reduced the number of parameters while retaining the important features, and techniques like dropout and earlystopping together to prevent overfitting.\\

At last, I have used some evaluation metrics to calculate the overall performance of my model.\\
The obtained results of evaluation metrics indicates that my model is somewhat good in capturing the trends required to predict the rise in stock prices but it lacks in capturing trends required to predict the fall in stock prices. I have also tried different ways to improve the performance of my model like by increasing or decreasing the number of layers in my model, by changing the kernel\_size of filters, also by using some weight initialization techniques but I could not achieve significant improvements in the figures of evaluation metrics that's why I decided to keep my present model as the final one.\\

There are some possible ways in which one can further lead a relevant study in order to improve the values of evaluation metrics to some extent:\\

\begin{enumerate}
    \item Although, I have trained my network on 1-D time series data using CONV1D function but one can also try to convert 1-D data into 2-D by creating plots of stock prices versus time and then can try to use these 2-D plots for training a CNN network using CONV2D function and then can compare the results.\\
    
    \item One can try to use some other features(if available) in the input data like the volume of stock prices per minute or volume weighted average price of stocks per minute and try to predict the movement of stock prices and then can compare the results.\\
    
    \item One can also try to fit a similar model on much larger datasets, this might enable the model to capture all the trends required to predict rise and fall in the stock prices with more efficacy.\\
    
    \item I have used only a CNN based model to predict the movement of stock prices; one can also try to construct models based on other deep learning algorithms like LSTM or both LSTM and CNN together to predict the movement of stock prices and then can compare the results.\\
    
\end{enumerate}

\section{References}
\begin{enumerate}
    \item Sheng Chen and Hongxiang He 2018,"Stock Prediction Using             Convolutional Neural Network". IOP Conf. Ser.: Mater. Sci. Eng. 435 012026.
    \item Yangqing Jia, Evan Shelhamer, Jeff Donahue, Sergey
          Karayev, Jonathan Long, Ross Girshick, Sergio
          Guadarrama, and Trevor Darrell. Caffe:"Convolutional architecture for fast feature embedding". arXiv
          preprint arXiv:1408.5093, 2014.
    \item Ashwin Siripurapu."Convolutional Networks for Stock                  Trading", 2015.
    \item J. F. Chen, W. L. Chen, C. P. Huang, S. H. Huang and A. P. Chen, “Financial Time-Series Data Analysis Using Deep  Convolutional Neural Networks,” 2016 7th International Conference on Cloud Computing and Big Data (CCBD), Macau, 2016, pp. 87-92.
    \item A. L. Maas, A. Y. Hannun, and A. Y. Ng. "Rectifier                 nonlinearities improve neural network 
          acoustic models". In ICML, 2013.
    \item Xu, Kelvin, et al. “Show, Attend and Tell: Neural Image           Caption Generation with Visual Attention.” Computer Science       (2015):2048-2057.
    \item Nelson, David M. Q., A. C. M. Pereira, and R. A. D.               Oliveira. “Stock market's price movement prediction with          LSTM neural networks.” International Joint Conference on          Neural Networks IEEE, 2017:1419-1426.
    \item Yann LeCun, Leon Bottou, Yoshua Bengio, and Patrick Haffner.
          "Gradient Based Learning Applied to Document Recognition",
          PROC. OF THE IEEE, NOVEMBER 1998.
    \item Yann LeCun, Leon Bottou, Genevieve B. Orr and Klaus-Robert        Muller. "Efficient Backprop".
    \item Yann LeCun, Yoshua Bengio. "Convolutional Networks for             Images, Speech and Time-Series".
    \item Ning Xue, Isaac Triguero, Grazziela P. Figueredo and Dario        Landa-Silva."Evolving Deep CNN-LSTMs for Inventory Time
          Series Prediction".
    \item John Gamboa, "Deep Learning for Time-Series Analysis"
    \item Sidra Mehtab, Jaydip Sen. "Stock Price Prediction Using CNN       and LSTM Based Deep Learning Models".
    \item Manuel R. Vargas, Beatriz S. L. P. de Lima and Alexandre G.       Evsukoff, "Deep learning for stock market prediction from         financial news articles",Conference: 2017 IEEE International       Conference on Computational Intelligence and Virtual              Environments for Measurement Systems and Applications             (CIVEMSA).
    \item X. Ding,  Y. Zhang,  T. Liu  and  J.  Duan,  “Using               Structured  Events  to  Predict Stock Price Movement: An          Empirical Investigation”, EMNLP, pp. 1415-1425, 2014.
    \item Y.  Kim, “Convolutional neural  networks  for  sentence          classification”,  arXiv preprint arXiv:1408.5882, 2014.
    \item Y. L. Boureau, F. Bach, Y. LeCun and J. Ponce, “Learning          mid-level features  for  recognition”,  IEEE  Conference  on       Computer  Vision  and  Pattern Recognition (CVPR), pp.            2559-2566, 2010.  
    \item H.  Mizuno,  M.Kosaka,  H.  Yajima  and  N.  Komoda,              “Application  of  neural  network  to  technical  analysis        of  stock  market  prediction”,  Studies in Informatic and        control, vol. 7, no. 3, pp. 111-120, 1998. 
    \item I. Parmar et al., "Stock Market Prediction Using Machine          Learning," 2018 First International Conference on Secure          Cyber Computing and Communication (ICSCCC), 2018, pp.             574-576, doi: 10.1109/ICSCCC.2018.8703332.
    \item Armano, G., M. Marchesi, and A. Murru. “A hybrid                  genetic-neural architecture for stock 
          indexes forecasting.” Information Sciences170.1(2005):3-33.
    \item K. Abhishek, A. Khairwa, T. Pratap and S. Prakash, “A stock       market prediction model using Artificial Neural Network,”         Computing Communication \& Networking Technologies (ICCCNT), 
          2012 Third International Conference on, Coimbatore, 2012, pp. 1-5.
    \item Diederik P. Kingma, Jimmy Ba, "Adam: A Method for Stochastic Optimization",  ICLR(2015).
    \item John C. Duchi, Elad Hazan, Y. Singer, "Adaptive Subgradient Methods for Online Learning and Stochastic Optimization"(2015).
    \item Bao Wang, Tan M. Nguyen, Andrea L. Bertozzi, Richard G. Baraniuk, Stanley J. Osher, "Scheduled Restart Momentum for Accelerated Stochastic Gradient Descent", arXiv:2002.10583v2.
    \item Ilya Sutskever, James Martens, George E. Dahl, Geoffrey E. Hinton,"On the importance of initialization and momentum in deep learning",  ICML(2013).
    \item Yann N. Dauphin, Harm de Vries, Junyoung Chung, Y. Bengio,
    "RMSProp and equilibrated adaptive learning rates for non-convex optimization", arXiv(2015).
    \item Xiaodong Cui, Wei Zhang, Zoltán Tüske and Michael Pichen, "Evolutionary Stochastic Gradient Descent for Optimization of Deep Neural Networks", 32nd Conference on Neural Information Processing Systems (NeurIPS 2018).
    \item Xue Ying 2019, "An Overview of Overfitting and its Solutions", J. Phys.: Conf. Ser. 1168 022022.
    \item Chiyuan Zhang, Oriol Vinyals, Remi Munos, Samy Bengio, "A Study on Overfitting in Deep Reinforcement Learning", arXiv:1804.06893v2.
    \item H. Li, J. Li, X. Guan, B. Liang, Y. Lai and X. Luo, "Research on Overfitting of Deep Learning," 2019 15th International Conference on Computational Intelligence and Security (CIS), 2019, pp. 78-81, doi: 10.1109/CIS.2019.00025.
    \item T. Wang, J. Huan and B. Li, "Data Dropout: Optimizing Training Data for Convolutional Neural Networks," 2018 IEEE 30th International Conference on Tools with Artificial Intelligence (ICTAI), 2018, pp. 39-46, doi: 10.1109/ICTAI.2018.00017.
    \item Z. Zhang, "Improved Adam Optimizer for Deep Neural Networks," 2018 IEEE/ACM 26th International Symposium on Quality of Service (IWQoS), 2018, pp. 1-2, doi: 10.1109/IWQoS.2018.8624183.\\

\end{enumerate}
\end{document}